\documentclass{article}
\usepackage{iclr2027_conference,times}

\usepackage{amsmath,amsfonts,bm}

\def\eqref#1{equation~\ref{#1}}

\def\1{\bm{1}}

\DeclareMathAlphabet{\mathsfit}{\encodingdefault}{\sfdefault}{m}{sl}
\SetMathAlphabet{\mathsfit}{bold}{\encodingdefault}{\sfdefault}{bx}{n}

\usepackage{amsmath,amssymb,booktabs,multirow,tabularx}
\usepackage{graphicx}
\usepackage{algorithm}
\usepackage{algpseudocode}
\usepackage{hyperref}
\usepackage{url}
\usepackage{wrapfig}
\usepackage{enumitem}

\title{Learning While Inferring: Local and Parallel Learning for Edge SNNs across Sensing Modalities}

\author{Yanxun Zhang$^{1,2*}$ \quad Yifei Wang$^{1,2}$\thanks{Equal Contribution. Correspondence to Xiaoqing Zheng (zhengxq@fudan.edu.cn).} \quad Changze Lv$^{1,2*}$ \quad Jingwen Xu$^{1,2}$ \quad Yiyang Lu$^{1,2}$ \quad \\ \textbf{Xiaohua Wang$^{1,2}$ \quad Di Yu$^3$ \quad Xin Du$^3$ \quad Xiaoqing Zheng$^{1,2}$} \\
$^1$ College of Computer Science and Artificial Intelligence, Fudan University\\
$^2$ Shanghai Key Laboratory of Intelligent Information Processing \\
$^3$ School of Software Technology, Zhejiang University\\
\texttt{\{yanxunzhang22,wangyif22,czlv24\}@m.fudan.edu.cn} \\
\texttt{\{zhengxq\}@fudan.edu.cn} \\
}

\newcommand{\stdv}[1]{{\scriptsize\textcolor{black!55}{$\pm$#1}}}

\iclrfinalcopy
\begin{document}
\maketitle

\begin{abstract}
Edge intelligence requires models to sense continuously in real time and to keep adapting on-device, all under tight compute, energy, and memory budgets. Although spiking neural networks (SNNs) enable efficient event-driven inference, standard surrogate-gradient backpropagation (BP) serializes updates and blocks ongoing inference. We investigate Bidirectional Spike-Based Distillation (BSD) as an on-device learning principle that lets edge SNNs learn while inferring. BSD couples a stimulus-driven forward network with an independent target-driven reverse network and aligns their intermediate representations through local objectives. Because the two pathways have disjoint computation graphs until alignment, forward inference, reverse inference, and stage-wise updates can run concurrently. On 25 benchmarks spanning the five sensing modalities of SOUL, BSD stays within 3.8 percentage points of matched BP baselines on average, while only its forward branch is needed at deployment. The learned representations also transfer well to few-shot class-incremental learning without replaying past data. By removing the dense floating-point backward chain, BSD reduces the projected training latency to $0.72\times$ and the estimated training energy to $0.36\times$ that of BP.
\end{abstract}

\section{Introduction}
\label{sec:introduction}

Edge computing brings intelligent processing directly to the sensors and end users that generate continuous data streams \citep{shi2016edge,zhou2019edge}. A defining trait of such deployments---whether in visual monitoring, keyword spotting, or wearable activity tracking---is that devices must sense continuously and in real time; they cannot simply stop watching or listening. Meanwhile, the operational environment is inherently non-stationary: after initial deployment, devices frequently encounter novel situations, such as a new user's voice or an unfamiliar spatial layout \citep{zhou2019edge,zhang2026safa}, and must therefore keep updating their models \citep{mcmahan2017communication,parisi2019continual}. Standard gradient-based training, however, imposes a strict sequential dependency: every forward pass must be followed by a complete global backward pass, which blocks the device from processing new inputs during the update \citep{di2024ec}. Integrating continual adaptation into always-on sensors therefore requires the ability to \textit{learn while inferring}, so that ongoing prediction and parameter updates overlap instead of blocking each other.

Spiking neural networks (SNNs) are particularly attractive for edge deployment because their sparse, event-driven communication and stateful temporal dynamics naturally suit sequential sensor streams \citep{roy2019towards, davies2018loihi}. However, although SNNs greatly reduce overall inference cost on neuromorphic hardware, their standard training methods are at odds with concurrent learning. Surrogate-gradient backpropagation (BP) propagates continuous floating-point errors through every layer and time step \citep{neftci2019surrogate,wu2018spatio,shrestha2018slayer}. This global dependency chain not only prevents inference during training but also incurs high latency and energy costs. Prior studies of on-device training show that naively ported, globally coupled training workloads are prohibitively slow and energy-intensive on battery-powered devices \citep{cai2020tinytl,lin2022device,zhu2023pockengine}. ANN-to-SNN conversion does not fundamentally avoid this problem either, since it adds a sequential two-stage pipeline that still relies on dense ANN computation \citep{deng2020rethinking}. Sparse spiking inference alone therefore cannot remove the blocking behavior and resource bottleneck of standard gradient-based training workloads.

\begin{figure*}[t]
  \vspace{-15pt}
  \centering
  \includegraphics[width=\textwidth]{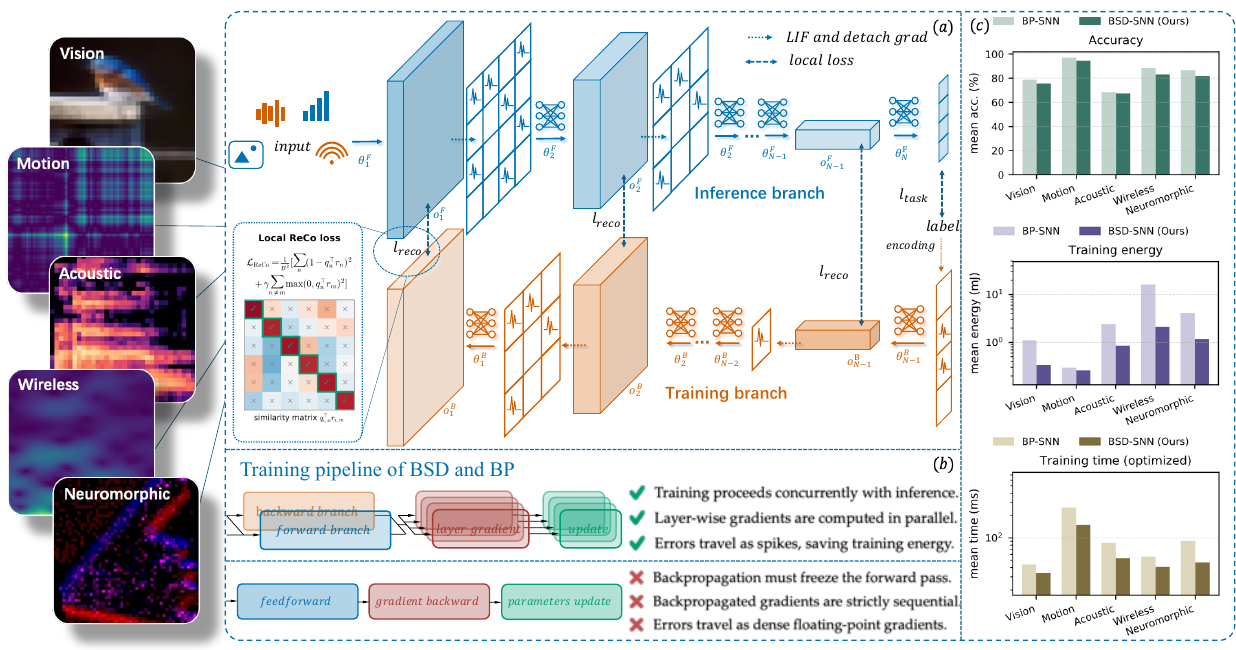}
  \caption{Overview of edge learning with BSD. Left: heterogeneous edge
  sensing modalities. (a) BSD couples a stimulus-driven forward
  (inference) branch with a target-driven reverse (training) branch aligned
  by local ReCo objectives. (b) The training pipeline of BSD and BP. BSD executes inference, reverse inference, and local
  updates concurrently, unlike the strictly sequential BP pipeline
  below. (c) BSD achieves accuracy comparable to BP across the five
  modalities while requiring substantially less training energy and a shorter
  projected training time.}
  \label{fig:teaser}
  \vspace{-11pt}
\end{figure*}

To enable learning while inferring, we studied Bidirectional Spike-Based Distillation (BSD) as a direct learning principle for edge SNNs. Originally proposed as a biologically plausible learning algorithm \citep{wang2026biologically}, BSD dispenses with the global error-backpropagation chain. Instead, it couples a stimulus-driven forward spiking network with an independent, target-driven reverse spiking network, and aligns their intermediate representations with local relaxed contrastive (ReCo) objectives \citep{lin2023relaxing} that provide layer-wise learning signals (Figure~\ref{fig:teaser}). Because the two branches run independently until their local losses are computed, neither depends on the other's intermediate outputs. This decoupling allows inference, reverse inference, and local updates to execute concurrently on parallel hardware, so that an edge SNN can learn while inferring.

Beyond concurrency, this independence also reduces training time and energy. By avoiding a floating-point error chain across all layers and time steps, BSD reduces the overall computational workload. Its layer-local gradient computations are mutually independent, which exposes parallelism and shortens the critical path of each training step. Moreover, replacing dense floating-point backpropagation with spike-based reverse inference substantially lowers the operation-level energy, bringing routine on-device adaptation within practical battery budgets.

Finally, the efficiency of local learning matters only if the learned representations remain strong. In dynamic edge scenarios, novel concepts typically arrive with only a few labels, and privacy constraints often preclude replaying historical data \citep{hersche2022constrained}. This setting is formalized as few-shot class-incremental learning (FSCIL) \citep{tao2020few}. Evaluating BSD under FSCIL, we show that even without global backpropagation, representations learned through local alignment remain discriminative and forward-compatible, supporting rapid adaptation to novel classes.

We evaluate BSD on the five edge-sensing modalities of the SOUL benchmark \citep{deng2025edge}: vision, motion, acoustic, wireless, and neuromorphic sensing. Our contributions are:
\begin{itemize}[noitemsep, topsep=0pt, leftmargin=*]
  \item We propose using BSD to enable \textit{learning while inferring} in edge SNNs, and show across multiple sensing modalities that its accuracy closely matches that of BP baselines.
  \item We characterize the resource advantages of BSD: branch-level and local-gradient parallelism yield a substantial projected reduction in training time, and spike-driven reverse inference yields a large reduction in operation-level training energy relative to BP.
  \item We show that BSD-trained representations support continual edge adaptation: In FSCIL, a BSD backbone learns novel classes from a few support examples through prototype-based adaptation.
\end{itemize}

\section{Related Work}
\label{sec:related}

\paragraph{Edge computing and online learning.}
Edge intelligence deploys inference and adaptation near data sources under tight computing, memory, and energy budgets \citep{shi2016edge,zhou2019edge}. Model compression and compact architectures reduce inference cost \citep{han2016eie,lin2020mcunet}, and neuromorphic hardware exploits event-driven dynamics for low-power temporal sensing \citep{roy2019towards,davies2018loihi}. Recently, the SOUL benchmark standardized cross-modal evaluation of edge SNNs \citep{deng2025edge}, and a modality-aware study showed that the efficiency benefits of SNNs vary across sensing modalities and hardware platforms \citep{du2026benchmarking}. Beyond static deployment, edge devices must adapt to novel concepts with scarce labels and without replaying past data, a challenge formalized by FSCIL \citep{tao2020few}. Prototype classifiers enable low-overhead class expansion \citep{snell2017prototypical}, while classifier evolution, semantic distillation, and forward-compatible representations mitigate catastrophic forgetting \citep{zhang2021few,cheraghian2021semantic,zhou2022forward}. C-FSCIL bounds the trainable memory footprint \citep{hersche2022constrained}, and SAFA-SNN performs sparse on-device adaptation of SNNs \citep{zhang2026safa}. These works make FSCIL a natural testbed for on-device adaptation under tight budgets.

\paragraph{Training spiking neural networks.}
Because SNN efficiency depends on event sparsity, simulation latency, and hardware support, low inference cost does not by itself imply efficient training \citep{deng2020rethinking}. Direct training with surrogate gradients assigns credit across layers and time steps \citep{neftci2019surrogate,wu2018spatio,shrestha2018slayer}, and later work reduces its latency and memory footprint \citep{meng2022training,meng2023towards}. Training cost has been further reduced through temporal gradient re-weighting \citep{deng2022temporal}, online updates that avoid storing the history \citep{xiao2022online,bellec2020solution}, and sparsity-promoting objectives \citep{kundu2021spike}. However, these methods remain tied to global backpropagation, whose cross-layer error chain serializes execution and blocks ongoing inference. Decoupled learning methods break this dependency through feedback alignment, direct feedback alignment, local error signals, synthetic gradients, and greedy layer-wise objectives \citep{lillicrap2016random,nokland2016direct,mostafa2018deep,jaderberg2017decoupled,belilovsky2019greedy}. For SNNs, DECOLLE learns from local readouts \citep{kaiser2020synaptic} and EC-SNN splits deep SNNs across edge devices \citep{di2024ec}; beyond SNNs, Counter-Current Learning trains two opposing networks \citep{kao2024counter}. BSD \citep{wang2026biologically} follows this direction with two opposing spiking pathways that are aligned locally at matched stages. Yet whether BSD is viable across edge sensing modalities, whether it supports concurrent inference and learning, and how well its representations generalize under few-shot incremental learning have not been systematically studied.

\section{Preliminaries}
\label{sec:preliminaries}

\subsection{Spiking Neural Networks}

We adopt the standard discrete-time leaky integrate-and-fire (LIF) neuron model \citep{maass1997networks}. Let $H^t$ and $U^t$ denote the membrane potential at time step $t$ immediately before and after the spike-triggered reset, respectively. The neuron dynamics are governed by
\begin{align}
  H^t &= U^{t-1}+\frac{1}{\tau}
  \left[I^t-\left(U^{t-1}-U_{\mathrm{reset}}\right)\right], \\
  S^t &= \mathbb{H}\!\left(H^t-U_{\mathrm{thr}}\right), \\
  U^t &= H^t(1-S^t)+U_{\mathrm{reset}}S^t,
\end{align}
where $I^t$ is the incoming external synaptic current, $\tau$ is the membrane time constant, $U_{\mathrm{thr}}$ is the firing threshold, $U_{\mathrm{reset}}=0$ is the reset potential, and $S^t\in\{0,1\}$ indicates the emitted discrete binary spike. Because the derivative of the Heaviside step function $\mathbb{H}(\cdot)$ is zero almost everywhere, we use surrogate-gradient learning \citep{neftci2019surrogate} and replace it with an arctangent (ATan) surrogate derivative with scale $\alpha=2$ in both the forward and reverse pathways. The explicit forms of $\mathbb{H}(\cdot)$ and the ATan surrogate derivative are given in Appendix~\ref{app:additional-preliminaries}.

\subsection{Few-Shot Class-Incremental Learning}
\label{sec:fscil-prelim}

Few-shot class-incremental learning (FSCIL) models settings in which the label space expands after deployment under tight supervision and storage constraints \citep{tao2020few}. Learning starts with a base session $\mathcal D^0$ over a class set $\mathcal C^0$, which is used to learn the initial representation. The system then encounters an ordered sequence of incremental sessions $\mathcal D^1,\ldots,\mathcal D^Q$. Each incremental session $q>0$ introduces a class set $\mathcal C^q$ that is disjoint from all previously observed classes ($\mathcal C^q\cap\mathcal C^j=\varnothing$ for $0\leq j<q$) and provides only $K$ labeled support samples for each of its $N$ new classes ($N$-way $K$-shot). After session $q$, the model must classify test samples over all classes seen so far, ${\mathcal C^{\leq q}=\bigcup_{j=0}^{q}\mathcal C^j}$, without retraining on data from earlier sessions. FSCIL therefore requires transferable base features, discriminative decision boundaries learned from very few samples, and resistance to catastrophic forgetting of earlier classes, mirroring edge deployments in which adaptation must proceed without replaying past data. After each session $q$, we report the seen accuracy on the test set $\mathcal D^{\mathrm{test}}_q$ restricted to $\mathcal C^{\le q}$ and, to expose any bias toward data-rich base classes over data-scarce novel ones, the harmonic mean of the per-class accuracies $a_c$ (H score):
\begin{equation}
  \mathrm{SeenAcc}_q=\frac{1}{|\mathcal D^{\mathrm{test}}_q|}
  \sum_{(x,y)\in\mathcal D^{\mathrm{test}}_q}\mathbb{I}[\hat y(x)=y],
  \qquad
  H_q=\frac{|\mathcal C^{\le q}|}{\sum_{c\in\mathcal C^{\le q}}a_c^{-1}}.
\end{equation}
Unlike the arithmetic mean, the harmonic mean is sensitive to poorly recognized classes, so a high H score requires balanced performance on both base and novel classes.

\section{Methods}
\label{sec:methods}

\subsection{Generalized BSD for Feedforward Models}
\label{sec:bsd}

Following \citet{wang2026biologically}, BSD decouples computation into two independent pathways: a stimulus-driven forward branch that predicts target labels from incoming sensory inputs, and a target-driven reverse branch that reconstructs inputs from spike-encoded labels. Because the two branches have separate computation graphs until alignment, forward inference never waits for the reverse branch, which enables concurrent inference and training.

Let $\mathbf a_i^t$ and $\mathbf b_i^t$ denote the binary spike tensors at the $i$-th hidden stage of the forward and reverse branches at time step $t$, respectively. The stage dynamics are
\begin{equation}
\mathbf a_i^t=\operatorname{LIF}\!\left(\tilde{\mathbf a}_i^t\right),
\qquad \tilde{\mathbf a}_i^t=f_i(\mathbf a_{i-1}^t;\theta_i^F),\qquad
\mathbf b_i^t=\operatorname{LIF}\!\left(\tilde{\mathbf b}_i^t\right),
\qquad \tilde{\mathbf b}_i^t=g_i(\mathbf b_{i+1}^t;\theta_i^B),
\end{equation}
where $f_i$ and $g_i$ are parameterized transformations, $\tilde{\mathbf a}_i^t$ and $\tilde{\mathbf b}_i^t$ are continuous pre-LIF activations, and the LIF function turns them into binary spikes. Here, $\mathbf a_0^t=x^t$ is the input sequence and $\mathbf b_N^t=e(y)^t$ is the spike-encoded target label. The two branches need not be symmetric, but they must expose the same ordered alignment stages $i=1,\ldots,N-1$.

The ReCo objective is computed on the continuous pre-LIF activations, which carry richer information than binary spikes. Let $q_{i,n}=\operatorname{norm}(\tilde{\mathbf a}_i^{1:T})_n$ and $r_{i,n}=\operatorname{norm}(\tilde{\mathbf b}_i^{1:T})_n$ denote the normalized features of sample $n$ in a batch of size $B$. The stage objective is
\begin{equation}
\mathcal L_{i}^{\mathrm{ReCo}}=\frac{1}{B^2}\left[
\sum_{n=1}^{B}(1-q_{i,n}^\top r_{i,n})^2+\gamma\sum_{n\ne m}
\big[\max(0,q_{i,n}^\top r_{i,m})\big]^2\right],
\end{equation}
where $\gamma$ weights the penalty on non-corresponding pairs. The task loss is applied only to the final forward readout. Following the layer-wise update rule of BSD, parameters of stage $i$ are updated as
\begin{align}
\theta_i^F&\leftarrow\theta_i^F-\eta_F\frac{\partial_{\mathrm{local}}}{\partial\theta_i^F}
\left[\lambda_i\mathcal L_i^{\mathrm{ReCo}}\right],\quad i=1,\ldots,N-1,\\
\theta_i^B&\leftarrow\theta_i^B-\eta_B\frac{\partial_{\mathrm{local}}}{\partial\theta_i^B}
\left[\lambda_i\mathcal L_i^{\mathrm{ReCo}}\right],\quad i=1,\ldots,N-1,\\
\theta_N^F&\leftarrow\theta_N^F-\eta_F
\frac{\partial\mathcal L_{\mathrm{task}}}{\partial\theta_N^F}.
\end{align}
The local derivative $\partial_{\mathrm{local}}$ treats the input to stage $i$ and the counterpart activations from the opposite branch as constants, so gradients stay within each stage and never cross layers.

\subsection{BSD for Temporal Models}
\label{sec:architectures}

For a input sequence $X=(x_1,\ldots,x_T)$ with one-hot target label $e_y$, temporal BSD runs the forward and reverse recurrences in opposite temporal directions. The forward RNN processes $X$ chronologically with spiking recurrent transitions $z_{t-1}=\operatorname{LIF}_F(h_{t-1})$:
\begin{equation}
h_t=W_{hh}^Fz_{t-1}+W_{ih}^Fx_t+b_h^F, \qquad
o_t=W_{ho}^Fh_t+b_o^F, \qquad
o=\frac{1}{T}\sum_{t=1}^{T}o_t.
\end{equation}
In contrast, the reverse RNN receives the constant one-hot target label $r_t=e_y$ at every step and runs sequentially backward in time from $T$ to $1$:
\begin{equation}
u_{t+1}=\operatorname{LIF}_B(q_{t+1}), \qquad
q_t=W_{hh}^Bu_{t+1}+W_{oh}^Br_t+b_h^B, \qquad
\hat{x}_t=W_{hi}^Bq_t+b_i^B.
\end{equation}
After the reverse hidden states are restored to chronological order, the overall training loss aligns the hidden states at every time step:
\begin{equation}
\mathcal L=\alpha\sum_{t=1}^{T}\mathcal L_{\mathrm{ReCo}}(h_t,q_t)+\beta\,\operatorname{CE}(o,y),
\end{equation}
where scalar hyperparameters $\alpha$ and $\beta$ balance the alignment and task terms. Multi-layer spiking LSTMs follow the same construction and align only the hidden states of every layer and time step, not the cell states. To control temporal credit assignment, the recurrent states can further be detached every $n$ steps, where $n=1$ gives a strictly time-local update and $n>1$ permits truncated backpropagation through time; details are given in Appendix~\ref{app:temporal-bsd}.

\subsection{Prototype-Based FSCIL}
\label{sec:fscilmethod}

To test whether representations learned by local alignment can accommodate label-space growth after deployment, we adopt the FSCIL protocol of Section~\ref{sec:fscil-prelim}. BSD first trains the SNN on all base-session data; the backbone is then frozen, and its forward features $z_l(x)=f_{\theta_F,l}(x)$ at layer $l$ are used directly to build class prototypes. For temporal inputs, $z_l(x)$ keeps its time dimension during the forward pass and is averaged over time only at extraction, $z_l(x)=T^{-1}\sum_t z_l^t(x)$. The prototype of class $c$ is the normalized mean feature of its support set $S_c$:
\begin{equation}
p_c=\operatorname{norm}\left(\frac{1}{|S_c|}\sum_{(x_i,y_i)\in S_c}
f_{\theta_F}(x_i)\right).
\end{equation}
In each incremental session $s>0$, novel prototypes are constructed only from the support set of $\mathcal C^s$. Base prototypes stay fixed and the backbone is not retrained, which avoids both catastrophic forgetting and computationally expensive updates. The classification score fuses cosine similarities from a set $\mathcal I$ of intermediate layers and adds a calibration bias:
\begin{equation}
S(x,c)=\sum_{l\in\mathcal I}\alpha_l\,\mathrm{cos}(z_l(x),p_{l,c})+b_c,
\qquad \sum_l\alpha_l=1,
\label{eq:proto-score}
\end{equation}
where $b_c$ is a pre-configured scalar that takes different values for base and novel classes, compensating for the systematic score gap between well-estimated base prototypes and data-scarce novel prototypes. The final discrete class prediction is obtained via
\begin{equation}
  \hat y=\arg\max_{c\in\mathcal C^{\le s}}S(x,c).
\end{equation}

To reduce the variance caused by the small support set, test-time augmentation (TTA) averages the extracted features of a set of augmented views $\mathcal V$ before scoring, i.e., $z_l(x)$ in Eq.~(\ref{eq:proto-score}) is replaced by $\bar z_l(x)=|\mathcal V|^{-1}\sum_{v\in\mathcal V}z_l(T_v(x))$. The same view averaging is used for both prototype construction and query inference, and all BSD backbone parameters remain unchanged. On-device adaptation thus reduces to forward feature extraction, prototype storage, and optional view averaging, all of which suit resource-constrained edge devices.

Following the subspace alignment idea of SAFA-SNN \citep{zhang2026safa}, we pull each novel prototype toward the subspace spanned by the base prototypes. Let $Q$ be an orthonormal basis of this subspace obtained by QR decomposition, and let $\alpha\in[0,1]$ be the projection strength. The calibrated novel prototype is
\begin{equation}
  p_c\leftarrow\operatorname{norm}\big((1-\alpha)\,p_c
  +\alpha\,QQ^{\top}p_c\big).
\end{equation}
This projection moves few-shot prototypes toward the region well covered by base classes, suppressing support-set noise while retaining their class-discriminative components.

\subsection{Time- and Energy-Aware BSD Execution}
\label{sec:analysis}

A BSD training step consists of four phases (Figure~\ref{fig:timeline}): forward inference ($T_F$), reverse-branch inference ($T_R$), local loss and gradient computation ($T_L$), and parameter update ($T_U$). The decoupled structure of BSD provides two sources of concurrency. First, because the forward and reverse branches have independent computation graphs until the ReCo objectives are formed, forward and reverse inference can run concurrently, so the branch latency on the critical path is $\max(T_F,T_R)$ rather than $T_F+T_R$. This property underlies \textit{learning while inferring}: an edge system can keep inferring on incoming sensor data while target-driven learning signals are computed alongside. Second, because each stage is trained by its own local loss (Section~\ref{sec:bsd}), the gradient computation of stage $i$ (taking time $T_{L,i}$) depends only on the representations recorded at alignment point $i$ and need not wait for error signals from later layers. Likewise, the disjoint parameter blocks can be updated independently (taking time $T_{U,i}$). In contrast, BP must propagate errors sequentially through the full network depth, which serializes the backward workload and yields a step latency of $T_F+T_L+T_U$.

With $P$ parallel workers assigned to stage-level tasks, local gradient computations and parameter updates can run concurrently, and the ideal BSD step latency becomes
\begin{equation}
  T_{\mathrm{BSD,opt}}=\max(T_F,T_R)+\frac{T_L+T_U}{P},
  \label{eq:bsd-latency}
\end{equation}
where $T_L=\sum_i T_{L,i}$ and $T_U=\sum_i T_{U,i}$ are the serial sums of the per-stage times, and $P$ is the number of parallel gradient tasks (the number of alignment points in feedforward networks; it scales with the sequence length in recurrent models). Realized speedups on edge hardware depend on the available parallel resources and on synchronization overhead. Nevertheless, in practical deployments where inference runs continuously and parameter updates arrive in small bursts, processing gradients asynchronously alongside forward inference can yield substantial efficiency gains.

Locality also reduces energy. BP propagates continuous floating-point errors through every layer and time step, which requires dense multiply-accumulate (MAC) operations amounting to roughly twice the dense forward workload. BSD replaces this backward chain with event-driven reverse inference, whose synaptic operations act on binary spikes and therefore reduce to accumulations (ACs), plus local gradient computations confined to each stage. This substitution is the main source of the training-energy savings quantified in Section~\ref{sec:exp-efficiency}.

\section{Experiments}
\label{sec:experiments}

\subsection{Dataset Processing}
\label{sec:exp-datasets}

We evaluate BSD on the SOUL edge-sensing benchmark \citep{deng2025edge}, which provides a unified, reproducible protocol for 25 tasks spanning five sensing modalities: vision (image classification), motion (activity recognition from inertial signals), acoustic sensing (environmental sound, keyword, and music-genre classification), wireless sensing (activity, identity, and gesture recognition from Wi-Fi channel state information, radar, and point clouds), and neuromorphic sensing (event-based vision and audio). The individual datasets are listed in Table~\ref{tab:main-results}.

\begin{table}[t]
\centering
\caption{Test accuracy (\%) of BP and BSD with matched forward architectures
(mean $\pm$ std, three runs). Models are convolutional SNNs unless marked
R (spiking RNN) or L (spiking LSTM).}
\label{tab:main-results}
\small
\setlength{\tabcolsep}{2pt}
\begin{tabular}[t]{@{}l@{\hspace{10pt}}r@{\,}l@{\hspace{10pt}}r@{\,}l@{}}
\toprule
Dataset & \multicolumn{2}{c}{BP} & \multicolumn{2}{c@{}}{BSD (Ours)} \\
\midrule
\multicolumn{5}{@{}l}{\textit{Vision}} \\
MNIST & 99.56 & \stdv{0.14} & 99.44 & \stdv{0.03} \\
Fashion-MNIST & 92.68 & \stdv{0.42} & 91.05 & \stdv{0.20} \\
CIFAR-10 & 87.12 & \stdv{1.76} & 84.13 & \stdv{0.34} \\
CIFAR-100 & 57.75 & \stdv{0.35} & 53.48 & \stdv{0.22} \\
SVHN & 94.36 & \stdv{0.73} & 90.81 & \stdv{0.11} \\
Tiny-ImageNet & 41.07 & \stdv{0.53} & 35.34 & \stdv{0.23} \\
\addlinespace[2pt]
\multicolumn{5}{@{}l}{\textit{Motion}} \\
UCI-HAR\textsuperscript{R} & 92.69 & \stdv{0.01} & 90.94 & \stdv{0.29} \\
HHAR\textsuperscript{L} & 98.54 & \stdv{0.12} & 96.37 & \stdv{0.05} \\
MotionSense\textsuperscript{R} & 98.95 & \stdv{0.58} & 96.30 & \stdv{0.06} \\
Shoaib\textsuperscript{L} & 98.17 & \stdv{0.06} & 94.86 & \stdv{0.29} \\
\addlinespace[2pt]
\multicolumn{5}{@{}l}{\textit{Acoustic}} \\
ESC-50 & 65.42 & \stdv{1.48} & 66.08 & \stdv{1.45} \\
UrbanSound8K & 77.68 & \stdv{0.69} & 74.15 & \stdv{1.11} \\
GSC\textsuperscript{L} & 92.33 & \stdv{0.37} & 84.03 & \stdv{0.88} \\
GTZAN & 62.67 & \stdv{0.62} & 62.17 & \stdv{1.25} \\
\bottomrule
\end{tabular}
\hspace{0.07\linewidth}
\begin{tabular}[t]{@{}l@{\hspace{10pt}}r@{\,}l@{\hspace{10pt}}r@{\,}l@{}}
\toprule
Dataset & \multicolumn{2}{c}{BP} & \multicolumn{2}{c@{}}{BSD (Ours)} \\
\midrule
\multicolumn{5}{@{}l}{\textit{Wireless}} \\
UT-HAR & 98.63 & \stdv{0.19} & 94.04 & \stdv{0.84} \\
NTU-Fi-HumanID & 93.47 & \stdv{0.31} & 92.37 & \stdv{0.99} \\
NTU-HAR & 99.24 & \stdv{0.31} & 98.48 & \stdv{0.62} \\
Widar3 & 78.51 & \stdv{0.41} & 65.82 & \stdv{0.32} \\
ARIL & 90.05 & \stdv{0.61} & 84.89 & \stdv{0.88} \\
BullyDetect & 69.23 & \stdv{2.31} & 60.54 & \stdv{1.22} \\
AOPHand & 84.72 & \stdv{1.54} & 80.09 & \stdv{0.19} \\
MMActivity & 94.59 & \stdv{0.04} & 90.29 & \stdv{0.25} \\
\addlinespace[2pt]
\multicolumn{5}{@{}l}{\textit{Neuromorphic}} \\
CIFAR10-DVS & 68.58 & \stdv{0.19} & 60.57 & \stdv{0.19} \\
DVS-Gesture & 86.46 & \stdv{0.57} & 83.45 & \stdv{0.82} \\
SHD & 99.66 & \stdv{0.05} & 98.33 & \stdv{0.09} \\
\midrule
\textbf{Average (25)} & \textbf{84.89} & & \textbf{81.12} & \\
\bottomrule
\end{tabular}
\vspace{-10pt}
\end{table}

To feed these heterogeneous raw streams to SNNs, we preprocess them into tensors of regular shape and scale. Images are normalized to $[0,1]$ and lightly augmented with standard spatial transforms. Motion signals are segmented into fixed windows (typically 128 time steps) that keep an explicit time axis for recurrent processing. Audio waveforms are converted into $64\times64$ single-channel log-mel spectrograms. Wireless measurements are standardized and reshaped into matrices that follow the native antenna layout or into voxel grids. Event streams are binned in time and rasterized into polarity-separated frame sequences. Each BSD model and its matched BP baseline consume exactly the same preprocessed tensors and train/test splits (including the base and incremental sessions in FSCIL), so the learning rule is the only experimental variable. Dataset details, tensor layouts, and modality-specific augmentations are given in Appendix~\ref{app:dataset-training}.

\subsection{Experimental Settings}
\label{sec:exp-settings}

We choose the model family according to the structure of each input. Convolutional SNNs process two-dimensional inputs (images, spectrograms, wireless tensors, and accumulated event frames), with the forward and reverse pathways exposing matched intermediate stages for alignment. Sequential tasks use recurrent SNNs: the BSD RNN for UCI-HAR and MotionSense, and a spiking LSTM with an LIF recurrent interface for HHAR, Shoaib, and Google Speech Commands (GSC). Every BP baseline uses the same forward architecture and input pipeline as its BSD counterpart. Architectural details are given in Appendix~\ref{app:dataset-training}.

To balance temporal fidelity and energy, we simulate $T=4$ time steps on all benchmarks, beyond which the original BSD study observed no clear accuracy gain \citep{wang2026biologically}, and use $T=10$ only for the event-camera datasets (CIFAR10-DVS and DVS-Gesture), whose streams span longer time windows. All LIF neurons use a membrane time constant $\tau=2.0$, a default firing threshold of $0.8$, and the ATan surrogate with $\alpha=2.0$. For the BSD objective, we set the ReCo penalty weight to $\gamma=0.6$, the layer-wise loss weight to $25.0$, and the task-loss weight to $1.0$. Both pathways are trained with AdamW, a cosine learning-rate schedule, gradient-norm clipping at $0.3$, and early stopping with a patience of 50 epochs. Dataset-specific settings, including batch sizes, learning rates, and weight decay, are listed in Table~\ref{tab:dataset-configs} (Appendix~\ref{app:dataset-training}). We evaluate classification accuracy, firing sparsity, training latency, and operation-level energy to profile BSD across modalities.

\subsection{Main Experimental Results}
\label{sec:exp-main}

Table~\ref{tab:main-results} compares BSD with the matched BP baselines on all 25 SOUL tasks. Across these datasets, BSD is competitive with BP, staying within $3.8$ percentage points of it on average ($81.12\%$ vs.\ $84.89\%$). The gap is small on most tasks: it is within $3$ points on 11 of the 25 benchmarks and within $5$ points on 19. BSD tracks BP most closely where the baseline accuracy is high; for example, it exceeds $90\%$ on all motion benchmarks. The remaining gap is concentrated on noisier or data-limited tasks such as Widar3, BullyDetect, and GSC.

The results on wireless tasks are partly affected by batch-size constraints. Because of the memory footprint of their large input tensors, several wireless benchmarks were trained with small minibatches of 4 to 16 samples. As analyzed in Appendix~\ref{app:batch-sensitivity}, the contrastive term of ReCo needs enough instance pairs per batch to shape the intermediate representations effectively. Since deployment uses only the forward spiking branch and BSD substantially reduces training latency and energy (Section~\ref{sec:exp-efficiency}), we regard this level of accuracy as a favorable trade-off for real-time edge adaptation.

\subsection{Training Time and Energy Analysis}
\label{sec:exp-efficiency}

\begin{figure}[t]
  \centering
  \includegraphics[width=\linewidth]{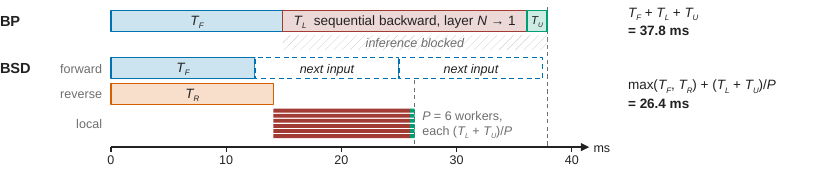}
  \caption{One CIFAR-10 training step drawn to scale (Table~\ref{tab:time-full}). BP
  runs its phases in sequence and blocks inference; BSD overlaps its branches
  and splits local work over $P=6$ workers.}
  \label{fig:timeline}
\end{figure}

\begin{figure}[t]
  \centering
  \includegraphics[width=\linewidth]{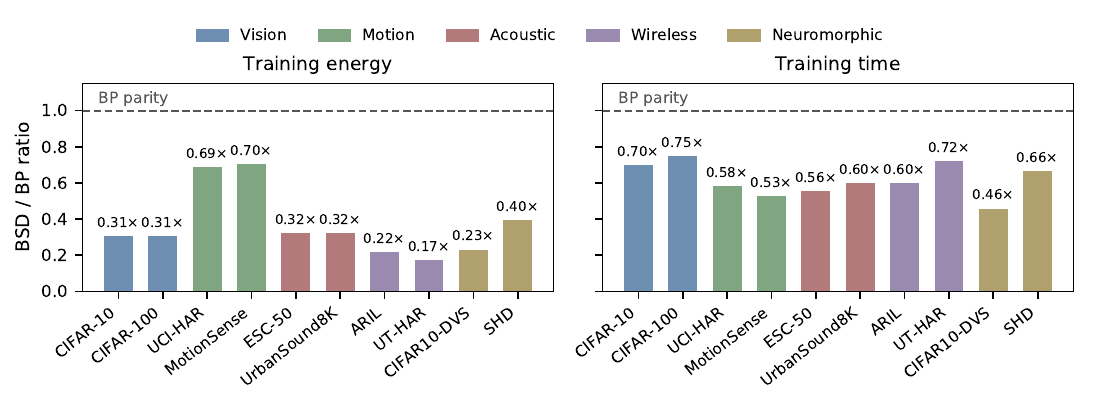}
  \caption{BSD/BP training-cost ratios on ten representative benchmarks (two per
  modality): operation-level energy (left) and projected time under concurrent
  execution (right); values below 1 mean that BSD is cheaper.}
  \label{fig:energy-time-ratio}
  \vspace{-8pt}
\end{figure}

Figure~\ref{fig:energy-time-ratio} compares the training cost of BSD and BP on ten representative benchmarks, two per modality. For latency, we profile the four phases of Section~\ref{sec:analysis} on a GPU and project $T_{\mathrm{BSD,opt}}$ with Eq.~(\ref{eq:bsd-latency}). For energy, we count the per-sample MACs on continuous values and ACs on binary spikes with operation hooks, and convert them to energy with the model in Appendix~\ref{app:energy-rules}. Per-phase timings, operation counts, and ratios for all 25 benchmarks are listed in Appendix~\ref{app:training-cost}.

When executed serially, a BSD iteration takes longer than a BP iteration on every dataset except NTU-Fi-HumanID, owing to the additional reverse-branch computation and the intermediate alignment losses. With the concurrent schedule of Section~\ref{sec:analysis} (Figure~\ref{fig:timeline}), however, the projected iteration latency falls below that of BP on 20 of the 25 benchmarks, with a geometric-mean ratio of $0.72\times$. The speedup depends on the available parallelism factor $P$. It is largest on neuromorphic and motion benchmarks (geometric means of $0.53\times$ and $0.62\times$, respectively): long simulation windows make the BP backward pass expensive on the former, and long recurrent horizons supply many independent gradient tasks on the latter. Conversely, shallow wireless models offer fewer concurrent stages; together with SVHN, which is essentially at parity ($1.004\times$), they account for the five benchmarks whose projected ratio exceeds 1. These projections assume ideal concurrency (Section~\ref{sec:analysis}).

The energy advantage is even more consistent. The training-energy ratio is below 1 on all 25 datasets, with a geometric mean of $0.36\times$. The reductions are largest in the wireless and neuromorphic domains (geometric means of $0.28\times$ and $0.30\times$, respectively), where high-dimensional inputs and sparse temporal firing maximize the benefit of spike-driven reverse execution. The margin is narrower on the recurrent motion benchmarks ($0.78\times$), where continuous recurrent state updates, including the LSTM gates, account for a large share of the arithmetic.

\subsection{FSCIL Performance}
\label{sec:exp-fscil}

\begin{wrapfigure}{r}{0.5\linewidth}
  \vspace{-10pt}
  \centering
  \includegraphics[width=\linewidth]{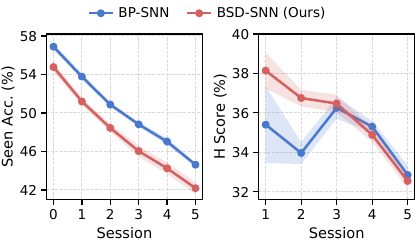}
  \vspace{-16pt}
  \caption{FSCIL on CIFAR-100: seen accuracy (left) and H score (right) per session, mean $\pm$ std over three runs.}
  \label{fig:fscil-trend}
  \vspace{-12pt}
\end{wrapfigure}
We next evaluate whether representations learned via local bidirectional distillation transfer effectively to FSCIL on CIFAR-100. Following Section~\ref{sec:fscilmethod}, the base session comprises 75 classes and is followed by five incremental sessions, each introducing 5 novel classes with 5 labeled support samples per class (5-way 5-shot). The base model is trained for 50 epochs to limit overfitting; subsequently, the forward backbone is frozen, and emerging classes are integrated exclusively via metric prototype construction without historical data replay (details in Appendix~\ref{app:fscil-config}). For comparison, an identical SNN is trained end-to-end with backpropagation. We report both cumulative seen accuracy and the harmonic mean ($H$ score) of per-class accuracies, where the latter penalizes performance disparity between base and novel categories.

Figure~\ref{fig:fscil-trend} plots both metrics across all sessions. BSD closely tracks the backpropagation baseline throughout. In the final session, BSD attains a seen accuracy of $42.18\%$ versus $44.62\%$ for backpropagation, alongside an $H$ score of $32.54\%$ versus $32.85\%$. Notably, BSD achieves slightly higher $H$ scores in the first two incremental sessions, indicating that local contrastive alignment maintains well-separated feature clusters. These results confirm that decentralized local training establishes robust, forward-compatible representations capable of continual post-deployment class expansion.

\subsection{Ablation Studies}
\label{sec:exp-ablations}

\begin{wraptable}{R}{0.46\linewidth}
  \vspace{-14pt}
  \centering
  \caption{Ablation of the adaptation components in the final FSCIL session (\%, mean $\pm$ std over three runs).}
  \label{tab:fscil-ablations}
  \small
  \setlength{\tabcolsep}{2.2pt}
  \begin{tabular}{@{}l r@{\,}l r@{\,}l@{}}
    \toprule
    Variant & \multicolumn{2}{c}{Seen Acc.} & \multicolumn{2}{c@{}}{H Score} \\
    \midrule
    Full pipeline & \textbf{42.18} & \stdv{0.60} & \textbf{32.54} & \stdv{0.36} \\
    w/o prototype TTA & 40.33 & \stdv{0.31} & 29.14 & \stdv{0.51} \\
    w/o multi-layer fusion & 38.47 & \stdv{0.12} & 31.86 & \stdv{0.28} \\
    w/o subspace proj. & 41.63 & \stdv{0.45} & 32.26 & \stdv{0.29} \\
    \bottomrule
  \end{tabular}
  \vspace{-10pt}
\end{wraptable}
We ablate the three key components of our prototype adaptation framework on CIFAR-100 (75 base classes, 5-way 5-shot): test-time view averaging (prototype TTA), multi-layer feature fusion, and base-subspace projection. All variants share the identical BSD-trained backbone and evaluation protocol.

As Table~\ref{tab:fscil-ablations} shows, removing any component degrades performance across both metrics. Multi-layer fusion has the largest impact on seen accuracy, which drops by $3.7$ percentage points when restricted to the penultimate layer alone, confirming that intermediate spiking representations encode complementary semantic hierarchies. Prototype TTA contributes most to the $H$ score ($+3.4$ points) by mitigating sampling noise in scarce five-shot support sets. Finally, base-subspace projection provides consistent regularization by anchoring novel prototypes toward the subspace of established base features. Operating without backbone retraining, these mechanisms provide an effective, training-free adaptation strategy for edge deployment.

\section{Conclusion}
\label{sec:conclusion}

This work studies BSD as an on-device learning principle for edge SNNs. To avoid the blocking behavior and resource cost of surrogate-gradient BP, BSD splits computation into a stimulus-driven forward branch and a target-driven reverse branch aligned by local objectives. With disjoint computation graphs until alignment, the two branches and their local updates can run concurrently, letting edge devices learn while inferring. On 25 benchmarks spanning five edge sensing modalities, BSD stays within $3.8$ percentage points of matched BP baselines on average while deploying only the forward branch, and its representations transfer well to 5-shot class-incremental learning. Branch- and stage-level parallelism reduces the projected step latency to $0.72\times$ that of BP, and the estimated training energy is lower on all 25 datasets (geometric mean $0.36\times$). These results position local bidirectional distillation as a promising foundation for always-on edge learning.

\section*{Reproducibility Statement}

The authors have made extensive efforts to ensure the reproducibility of the
empirical results reported in this paper. First, detailed dataset
characteristics and preprocessing procedures are documented in
Appendix~\ref{app:dataset-training}. Second, comprehensive implementation
details, covering network architectures, hyperparameter settings, training
procedures, and evaluation protocols, are provided in
Appendices~\ref{app:dataset-training} and~\ref{app:training-cost}. All accuracy
results are averaged over three runs, and the source code, configurations, and
plotting scripts will be released upon publication.

\section*{Ethics Statement}

Our experiments use public benchmark datasets spanning human activity,
identity, audio, wireless, and event-camera sensing; each dataset was
collected under the protocols of its original study, and this work collects no
new human data. Wireless and camera-free sensing can enable monitoring or
identification without a subject's awareness, so applications built on such
models should be deployed only with informed consent and appropriate access
control. On-device processing reduces the need to transmit raw sensor data,
but does not by itself guarantee privacy or prevent misuse. Finally, our
efficiency figures are operation-level estimates and projections rather than
measurements on specific hardware, and should be interpreted as such.

\bibliography{references}
\bibliographystyle{iclr2027_conference}

\newpage

\appendix
\section{Additional Preliminaries}
\label{app:additional-preliminaries}

\subsection{Surrogate Gradient and Heaviside Step Function}

In Section~\ref{sec:preliminaries}, the discrete firing activity of the LIF neuron is determined by the Heaviside step function $\mathbb{H}(x)$, defined as
\begin{equation}
  \mathbb{H}(x)=
  \begin{cases}
    1, & x\geq 0,\\
    0, & x<0.
  \end{cases}
\end{equation}
The discontinuous thresholding step results in a derivative $\mathbb{H}'(x)$ that is zero almost everywhere and undefined at $x=0$, preventing informative gradient backpropagation through discrete spike events. 

Surrogate gradient learning circumvents this issue by introducing a smooth approximation only during backward gradient computation \citep{neftci2019surrogate,wu2018spatio,shrestha2018slayer}. In our experiments, we adopt the arctangent (ATan) surrogate. Its smooth primitive function $\widetilde{\mathbb H}_{\alpha}(x)$ and the corresponding surrogate derivative are
\begin{align}
  \widetilde{\mathbb H}_{\alpha}(x)
  &= \frac{1}{\pi}\arctan\!\left(\frac{\pi}{2}\alpha x\right)+\frac{1}{2},\\
  \frac{\partial \widetilde{\mathbb H}_{\alpha}(x)}{\partial x}
  &= \frac{\alpha}{2\left[1+\left(\frac{\pi}{2}\alpha x\right)^2\right]},
\end{align}
where $\alpha>0$ controls the steepness of the derivative around the firing threshold. We use $\alpha=2$ for all models. Activations remain strictly binary during the forward simulation, while non-zero surrogate gradients arise near $H^t=U_{\mathrm{thr}}$.

\section{Detailed BSD Objectives and Pseudocode}
\label{app:bsd-pseudocode}

\subsection{Temporal BSD with Spiking LSTMs}
\label{app:temporal-bsd}

\begin{figure}[ht]
  \centering
  \includegraphics[width=\linewidth]{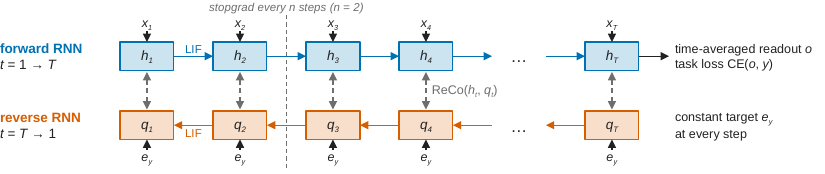}
  \caption{Temporal BSD (Section~\ref{sec:architectures}) unrolled over time.
  The forward RNN runs from $t=1$ to $T$ on the inputs $x_t$, the reverse RNN
  runs from $t=T$ to $1$ on the constant target $e_y$, and ReCo aligns their
  hidden states $h_t$ and $q_t$ at every step; the task loss acts only on the
  time-averaged forward readout. Detaching the recurrent states every $n$
  steps limits temporal credit assignment to segments of length $n$.}
  \label{fig:temporal}
\end{figure}

Figure~\ref{fig:temporal} illustrates the temporal construction of
Section~\ref{sec:architectures}. For an $M$-layer spiking LSTM with hidden states $h_t$ and cell states $c_t$,
the reverse LSTM likewise traverses the sequence backward in time, and ReCo
alignment is applied to the hidden states $h_t^{(m)}$ and $\bar h_t^{(m)}$ of
the two LSTMs at every layer and time step:
\begin{equation}
\mathcal L=\alpha\sum_{t=1}^{T}\sum_{m=1}^{M}
\mathcal L_{\mathrm{ReCo}}(h_t^{(m)},\bar h_t^{(m)})
+\beta\,\operatorname{CE}(p,y),\qquad
p=\frac{1}{T}\sum_{t=1}^{T}(W_{\mathrm{cls}}h_t+b_{\mathrm{cls}}).
\end{equation}
Cell states act as internal memory and are not aligned.

To control temporal credit assignment, we split each sequence into segments of
length $n$ and detach the recurrent states only at segment boundaries,
\begin{equation}
(h_{kn}^{(m)},c_{kn}^{(m)})\leftarrow
\operatorname{stopgrad}(h_{kn}^{(m)},c_{kn}^{(m)}),
\end{equation}
so that gradients propagate only within a segment. Setting $n=1$ gives a
strictly time-local update, whereas $n>1$ permits truncated backpropagation
through time within each segment without changing the alignment points.

\subsection{Pseudocode}

Algorithms~\ref{alg:parallel-bsd} and~\ref{alg:bsd-branches} give the complete
BSD training step for feedforward (CNN) and temporal (RNN/LSTM) models,
respectively. In both cases the two branches run concurrently, every aligned
representation is detached after being recorded so that gradients never cross
an alignment point, and the per-stage gradient tasks and parameter updates are
independent and therefore parallelizable.

\begin{algorithm}[H]
\caption{BSD training step for feedforward (CNN) models}
\label{alg:parallel-bsd}
\begin{algorithmic}[1]
\Require minibatch $(x,y)$; forward blocks $\{f_i\}_{i=1}^{N}$ with parameters
$\theta^F$; backward blocks $\{g_i\}_{i=1}^{N}$ with parameters $\theta^B$;
alignment points $\mathcal A\subseteq\{1,\ldots,N-1\}$
\State $\mathbf a_0 \gets x$ (spike-encoded input);\quad
       $\mathbf b_{N+1} \gets e(y)$ (spike-encoded label)
\Statex \hspace{-1em}\textbf{run concurrently:}
\State \textbf{forward:} \textbf{for} $i=1,\ldots,N$ \textbf{do}
\Statex \hspace{1em}$\tilde{\mathbf a}_i \gets f_i(\mathbf a_{i-1})$;\quad
       if $i\in\mathcal A$: record $\tilde{\mathbf a}_i$ and
       $\mathbf a_i \gets \operatorname{Detach}\!\big(\operatorname{LIF}(\tilde{\mathbf a}_i)\big)$;\quad
       else $\mathbf a_i \gets \operatorname{LIF}(\tilde{\mathbf a}_i)$
\State \textbf{backward:} \textbf{for} $i=N,\ldots,1$ \textbf{do}
\Statex \hspace{1em}$\tilde{\mathbf b}_i \gets g_i(\mathbf b_{i+1})$;\quad
       if $i\in\mathcal A$: record $\tilde{\mathbf b}_i$ and
       $\mathbf b_i \gets \operatorname{Detach}\!\big(\operatorname{LIF}(\tilde{\mathbf b}_i)\big)$;\quad
       else $\mathbf b_i \gets \operatorname{LIF}(\tilde{\mathbf b}_i)$
\State $\mathcal L_{\mathrm{task}} \gets
       \operatorname{CE}\big(\operatorname{Readout}(\mathbf a_N),\,y\big)$;\quad
       $g_N^F \gets \nabla_{\theta_N^F}\mathcal L_{\mathrm{task}}$
\State \textbf{in parallel for $i\in\mathcal A$:}
\Statex \hspace{1em}$g_i^F \gets \nabla_{\theta_i^F}\,
       \lambda_i\mathcal L_{\mathrm{ReCo}}\big(\tilde{\mathbf a}_i,\,
       \operatorname{Detach}(\tilde{\mathbf b}_i)\big)$
\Statex \hspace{1em}$g_i^B \gets \nabla_{\theta_i^B}\,
       \lambda_i\mathcal L_{\mathrm{ReCo}}\big(
       \operatorname{Detach}(\tilde{\mathbf a}_i),\,\tilde{\mathbf b}_i\big)$
\State Synchronize, then \textbf{in parallel} update every disjoint parameter
       block $\theta_i^F,\theta_i^B$ with its gradient
\end{algorithmic}
\end{algorithm}
\vspace{-10pt}
\begin{algorithm}[H]
\caption{BSD training step for temporal (RNN/LSTM) models}
\label{alg:bsd-branches}
\begin{algorithmic}[1]
\Require sequence $x_{1:T}$; label $y$; forward and backward recurrent cells
$\operatorname{RecCell}_F,\operatorname{RecCell}_B$ (RNN or LSTM) with
parameters $\theta^F,\theta^B$; segment length $n$
\State $h_0 \gets 0$;\quad $q_{T+1} \gets 0$
\Statex \hspace{-1em}\textbf{run concurrently:}
\State \textbf{forward:} \textbf{for} $t=1,\ldots,T$ \textbf{do}
\Statex \hspace{1em}$h_t \gets \operatorname{RecCell}_F(x_t, h_{t-1})$;\quad
       if $t \bmod n = 0$: $h_t \gets \operatorname{Detach}(h_t)$
\State \textbf{backward:} \textbf{for} $t=T,\ldots,1$ \textbf{do}
\Statex \hspace{1em}$q_t \gets \operatorname{RecCell}_B\big(e(y), q_{t+1}\big)$;\quad
       if $t \bmod n = 0$: $q_t \gets \operatorname{Detach}(q_t)$
\State Restore $\{q_t\}$ to chronological order
\State $o \gets \frac{1}{T}\sum_{t=1}^{T} W_{ho} h_t$;\quad
       $\mathcal L_{\mathrm{task}} \gets \operatorname{CE}(o, y)$
\State \textbf{in parallel for $t=1,\ldots,T$ (and each layer $m$):}
\Statex \hspace{1em}$\hat g_t^F \gets \nabla_{\theta^F}\,
       \alpha\,\mathcal L_{\mathrm{ReCo}}\big(h_t,\,
       \operatorname{Detach}(q_t)\big)$
\Statex \hspace{1em}$\hat g_t^B \gets \nabla_{\theta^B}\,
       \alpha\,\mathcal L_{\mathrm{ReCo}}\big(
       \operatorname{Detach}(h_t),\,q_t\big)$
\State Synchronize and reduce gradients of the time-shared weights:
       $g^F \gets \sum_{t=1}^{T}\hat g_t^F + \nabla_{\theta^F}\mathcal L_{\mathrm{task}}$;\quad
       $g^B \gets \sum_{t=1}^{T}\hat g_t^B$
\State Update the disjoint parameter blocks (layers, branches) in parallel
\Statex \hspace{-1em}\emph{LSTM cells additionally carry a cell state $c_t$,
which is detached at segment boundaries together with $h_t$ but never serves
as a ReCo alignment target.}
\end{algorithmic}
\end{algorithm}

\section{Dataset and Training Details}
\label{app:dataset-training}

\subsection{Dataset Composition and Preprocessing}

\paragraph{Vision.}
MNIST and Fashion-MNIST are loaded through torchvision and resized to
$32\times32$ single-channel images. CIFAR-10, CIFAR-100, and SVHN retain their
native $32\times32$ RGB format with the standard train/test partitions, and
Tiny-ImageNet is used at its native $64\times64$ resolution. All image tensors are normalized to $[0,1]$, and light standard
augmentation is applied during training.

\paragraph{Motion.}
UCI-HAR provides nine-channel inertial signals segmented into windows of length
$128$ with the official train/test split, yielding $[128,9]$ sequences for the
recurrent models. HHAR uses accelerometer and gyroscope readings from the
official CSV logs; records without labels are discarded, windows of length
$128$ and stride $64$ are extracted, and one held-out user forms the test set,
giving $[128,3]$ sequences. MotionSense provides twelve device-motion features
per step and uses the same windowing, with subjects 21--24 forming the test set
($[128,12]$). Shoaib is windowed identically, with participants 9 and 10 held
out for testing.

\paragraph{Acoustic and Speech.}
ESC-50 (50 environmental sound classes), UrbanSound8K (urban sound events,
official folds 9 and 10 held out for testing), GSC (keyword spotting), and
GTZAN (music genres) are stored as waveform manifests at the preprocessing
stage; each waveform is converted to a $64\times64$ single-channel log-mel
spectrogram when loaded, so all acoustic tasks share a common image-like input
format.

\paragraph{Wireless.}
UT-HAR channel state information (CSI) samples are reshaped to $[1,250,90]$, with the training and
validation partitions merged for training. NTU-Fi-HumanID and NTU-HAR
compress and rearrange the raw $342\times2000$ CSI amplitude into
$[3,114,500]$ tensors with fixed standardization. Widar3 standardizes the
$22\times400$ CSI features and reshapes them to $[22,32,32]$. ARIL uses the
amplitude field only, giving $[1,52,192]$ matrices. BullyDetect reshapes the
$90\times T$ CSI stream to $3\times30\times T$ and interpolates it to
$[3,32,500]$. AOPHand reads per-gesture CSV feature sequences with an 80/20
per-class split and pads them to a common two-dimensional size (or keeps the
$[T,F]$ sequence for the recurrent variant). MMActivity voxelizes the RadHAR
point clouds into $10\times32\times32$ volumes over 60-frame windows.

\paragraph{Neuromorphic Sensing.}
CIFAR10-DVS events are accumulated into ten frames at $48\times48$ resolution
with two polarity channels. DVS-Gesture follows the official trial split and is
accumulated into ten frames at $64\times64$. SHD spike trains are rasterized
from the official HDF5 files into $[1,64,64]$ tensors.

\subsection{Model Architectures}

\paragraph{Feedforward models.}
Image, spectrogram, event-frame, and matrix-shaped inputs are processed by a
convolutional SNN whose blocks comprise convolution, optional batch
normalization, LIF dynamics, and $2\times2$ pooling, followed by flattening and
a linear readout. Channel widths, pooling locations, and kernel sizes are
adapted to each input resolution (e.g., channels $[128,256,256]$ for
CIFAR10-DVS). The BSD reverse branch mirrors the forward block structure with
nearest-neighbor unpooling and reverse convolutions, and the two branches are
aligned at every block output.

\paragraph{Temporal models.}
Sequence tasks use the BSD RNN (a single recurrent layer with independent
forward input/hidden/readout weights and reverse output/hidden/input weights),
the spiking LSTM (whose analog hidden state passes through an LIF neuron so
that the recurrent signal is spiking), or a GRU variant, all sharing the same
forward/reverse alignment interface. Hidden states are aligned at every time
step between the two branches, as defined in Section~\ref{sec:architectures} and
Appendix~\ref{app:temporal-bsd}.

\paragraph{BSD and BP correspondence.}
Every BP baseline reuses exactly the forward architecture of its BSD
counterpart---same layer widths, kernels, pooling, and readout---and is trained
end-to-end with the same surrogate gradient. BSD adds the label-driven reverse
branch, which is used only during training and discarded at deployment, so the
comparison isolates the effect of the learning rule.

\subsection{Training Configurations}

\paragraph{Common training settings.}
Unless stated otherwise, all models are trained with AdamW
($\beta_1=0.9$, $\beta_2=0.95$, $\epsilon=10^{-8}$), a cosine learning-rate
schedule with a 30-epoch warm-up over a 120-epoch horizon, at most 200 epochs,
and early stopping with patience 50; gradients of both parameter sets are
clipped to a maximum norm of $0.3$. SNNs simulate $T=4$ time steps ($T=10$ for
CIFAR10-DVS and DVS-Gesture) with membrane constant $\tau=2.0$, firing
threshold $0.8$, hard reset, and the ATan surrogate with slope $\alpha=2.0$.
BSD uses ReCo penalty weight $\gamma=0.6$, layer-loss weight $25.0$,
task-loss weight $1.0$, and numerical constant $10^{-8}$; batch normalization
has affine parameters in both branches. Image inputs are normalized to $[0,1]$
with standard light augmentation; audio and wireless experiments keep the SOUL
augmentations; neuromorphic inputs additionally use event drop ($p=0.05$),
temporal drop (ratio $0.1$), and spatial shifts of up to $2$ pixels. Every
configuration is run with three seeds.
\vspace{-4pt}
\paragraph{Dataset-specific settings.}
Table~\ref{tab:dataset-configs} reports the model family, batch size, learning
rate, weight decay, and firing thresholds for each dataset. The forward and
reverse parameter sets always share the same optimizer settings (learning rate,
weight decay, and gradient clipping); only the firing threshold differs between
the two branches, on two datasets.

\begin{table*}[h]
  \vspace{-16pt}
  \centering
  \footnotesize
  \setlength{\tabcolsep}{4pt}
  \caption{Per-dataset training configurations. $U_{\mathrm{thr}}^{F}$ and
  $U_{\mathrm{thr}}^{B}$ are the firing thresholds of the forward and reverse
  branches. Rows marked $^\dagger$ follow the configurations of the original
  BSD paper \citep{wang2026biologically}.}
  \label{tab:dataset-configs}
  \begin{tabular}{@{}lllcccc@{}}
    \toprule
    Modality & Dataset & Model & Batch size & Learning rate & Weight decay
    & $U_{\mathrm{thr}}^{F}$\,/\,$U_{\mathrm{thr}}^{B}$ \\
    \midrule
    \multirow{6}{*}{Vision}
    & MNIST$^\dagger$ & CNN & 128 & $10^{-4}$ & $10^{-5}$ & 0.8/0.8 \\
    & Fashion-MNIST$^\dagger$ & CNN & 128 & $10^{-4}$ & $10^{-5}$ & 0.8/0.8 \\
    & CIFAR-10$^\dagger$ & CNN & 128 & $10^{-4}$ & $10^{-5}$ & 0.8/0.8 \\
    & CIFAR-100$^\dagger$ & CNN & 128 & $10^{-4}$ & $10^{-5}$ & 0.8/0.8 \\
    & SVHN$^\dagger$ & CNN & 128 & $10^{-4}$ & $10^{-5}$ & 0.8/0.8 \\
    & Tiny-ImageNet$^\dagger$ & CNN & 128 & $10^{-4}$ & $10^{-5}$ & 0.8/0.8 \\
    \midrule
    \multirow{4}{*}{Motion}
    & UCI-HAR & RNN & 128 & $10^{-3}$ & $10^{-6}$ & 0.8/0.8 \\
    & HHAR & Spiking LSTM & 128 & $10^{-3}$ & $10^{-6}$ & 0.8/0.8 \\
    & MotionSense & RNN & 128 & $10^{-3}$ & $10^{-6}$ & 0.8/0.8 \\
    & Shoaib & Spiking LSTM & 128 & $10^{-3}$ & $10^{-6}$ & 0.8/0.8 \\
    \midrule
    \multirow{4}{*}{Acoustic}
    & ESC-50 & CNN & 32 & $10^{-3}$ & $10^{-6}$ & 0.8/0.8 \\
    & UrbanSound8K & CNN & 40 & $10^{-3}$ & $10^{-6}$ & 0.8/0.8 \\
    & GSC & Spiking LSTM & 128 & $10^{-3}$ & $10^{-6}$ & 0.6/0.6 \\
    & GTZAN & CNN & 32 & $10^{-3}$ & $10^{-6}$ & 0.8/0.8 \\
    \midrule
    \multirow{8}{*}{Wireless}
    & UT-HAR & CNN & 14 & $10^{-3}$ & $10^{-6}$ & 0.8/0.8 \\
    & NTU-Fi-HumanID & CNN & 11 & $10^{-3}$ & $10^{-6}$ & 0.8/0.8 \\
    & NTU-HAR & CNN & 4 & $10^{-3}$ & $10^{-6}$ & 0.8/0.8 \\
    & Widar3 & CNN & 32 & $10^{-3}$ & $10^{-6}$ & 0.8/0.8 \\
    & ARIL & CNN & 32 & $10^{-3}$ & $10^{-6}$ & 0.8/0.8 \\
    & BullyDetect & CNN & 16 & $3\times10^{-4}$ & $10^{-6}$ & 0.8/0.8 \\
    & AOPHand & WiFi-CNN & 128 & $10^{-3}$ & $10^{-6}$ & 0.6/0.8 \\
    & MMActivity & WiFi-CNN & 64 & $10^{-3}$ & $10^{-6}$ & 0.8/0.8 \\
    \midrule
    \multirow{3}{*}{Neuromorphic}
    & CIFAR10-DVS & CNN & 25 & $10^{-4}$ & $10^{-6}$ & 0.8/0.8 \\
    & DVS-Gesture & CNN & 12 & $3\times10^{-4}$ & $10^{-6}$ & 0.6/0.8 \\
    & SHD & CNN & 50 & $10^{-3}$ & $10^{-6}$ & 0.8/0.8 \\
    \bottomrule
  \end{tabular}
  \vspace{-20pt}
\end{table*}
\subsection{FSCIL Configuration}
\label{app:fscil-config}

\begin{figure}[h]
  \centering
  \includegraphics[width=\linewidth]{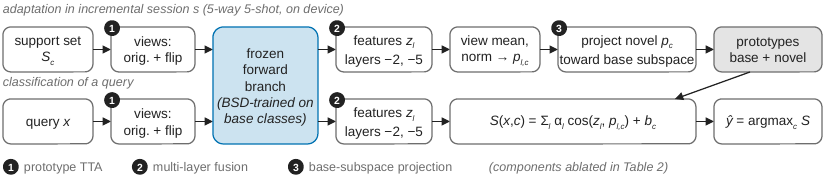}
  \caption{Prototype-based FSCIL on the frozen BSD forward branch
  (Section~\ref{sec:fscilmethod}). In each incremental session, novel
  prototypes are built from view-averaged support features at two layers and
  projected toward the subspace of the base prototypes; a query is classified
  by its fused cosine scores against all prototypes. Numbered markers indicate
  the components ablated in Table~\ref{tab:fscil-ablations}.}
  \label{fig:fscil-pipeline}
\end{figure}

FSCIL experiments use CIFAR-100 with a fixed class order (split seed $2025$):
session $0$ contains $75$ base classes, and each of the five incremental
sessions introduces $5$ novel classes ($5$-way) with $5$ labeled support
examples per class ($5$-shot), sampled with support seed $2025$. The base
session trains the BSD CNN for $50$ epochs on all base-class training samples;
afterwards the forward backbone is frozen. Prototypes are compared with cosine
similarity, and the scores of two feature levels ($-2$ and $-5$) are fused with
weights $0.85$ and $0.15$. Novel prototypes are built from view-averaged support
features (the original and horizontally flipped views), optionally evaluated
with the same view averaging at test time, and projected toward the subspace
spanned by the base prototypes with strength $\alpha=0.3$, following the
prototype-space projection of SAFA-SNN \citep{zhang2026safa}.
Figure~\ref{fig:fscil-pipeline} summarizes the resulting pipeline.

\section{Training Cost Measurements}
\label{app:training-cost}

\subsection{Energy Counting Rules}
\label{app:energy-rules}

Our energy estimates distinguish two classes of arithmetic. When the input of
a layer is a binary spike train, each synaptic multiplication degenerates into
an accumulation and is counted as one AC; when the input is continuous-valued
(e.g., raw sensor inputs or pre-LIF activations), the operation is counted as
one multiply--accumulate (MAC). Following the 45\,nm operation-level model of
\citet{horowitz20141}, we assign $E_{\mathrm{MAC}}=4.6$ pJ and
$E_{\mathrm{AC}}=0.9$ pJ. All SNN time steps are included through the
multi-step tensor shapes. For spike-driven layers, we measure the mean firing
rate of the actual binary input on the recorded batch and count ACs as the
layer's dense synaptic operations scaled by this rate, so no firing-rate
assumption is needed.

For BP, we count the forward pass layer by layer using operation hooks on
every convolution, linear, and recurrent-cell call. The backward pass is
estimated as twice the forward dense workload, accounting for the weight
gradient and the input gradient of every layer; this global backward chain is
the dominant training cost of BP.

For BSD, the forward and reverse branches are each counted with the same rule.
The reverse branch is driven by the spike-encoded label, so most of its
operations are ACs, which is one source of BSD's energy advantage. Because each
hidden level is updated only by its own local objective, the backward
computation of a level stays within that level and is counted once per level;
there is no cross-layer error chain. The ReCo objective adds the cost of its
similarity matrices: at each alignment point we count $B^{2}D$ MACs for the
forward similarities, with $B$ the batch size and $D$ the representation
dimension, and twice that for their backward pass.

Summing the two operation classes over all counted components gives the
per-sample energy of one training step,
\begin{equation}
  E_{\mathrm{train}}
  =E_{\mathrm{MAC}}\times N_{\mathrm{MAC}}
  +E_{\mathrm{AC}}\times N_{\mathrm{AC}}
  =4.6\,\mathrm{pJ}\times N_{\mathrm{MAC}}
  +0.9\,\mathrm{pJ}\times N_{\mathrm{AC}},
\end{equation}
which is the energy model used in Section~\ref{sec:exp-efficiency};
Table~\ref{tab:energy-full} reports the resulting $N_{\mathrm{MAC}}$ and
$N_{\mathrm{AC}}$ for every dataset. These estimates are analytic,
operation-level quantities: the constants come from a 45\,nm literature model
rather than measurements on a specific device, so the values should be read as
a hardware-independent proxy for training cost, not as measured energy on edge
or neuromorphic hardware.

\subsection{Per-Dataset Measurements}

Tables~\ref{tab:energy-full} and~\ref{tab:time-full} report the complete
per-dataset measurements behind Figure~\ref{fig:energy-time-ratio}.
Table~\ref{tab:energy-full} lists operation counts and estimated training
energy. MACs are multiply--accumulate operations on continuous-valued
activations; ACs are accumulation-only operations on binary spike inputs; both
are counted per sample over a full training step (forward, reverse, and local
gradient computations; BP backward is counted as twice the forward dense
workload). Energy follows the operation-level model of
\citet{horowitz20141} ($4.6$ pJ per MAC and $0.9$ pJ per AC).
Table~\ref{tab:time-full} lists the profiled stage times per training step
(forward $T_F$, reverse $T_R$, gradient computation $T_L$, parameter update
$T_U$) together with the optimized BSD time
$T_{\mathrm{BSD,opt}}=\max(T_F,T_R)+(T_L+T_U)/P$, where $P$ is the
gradient-parallelism factor (number of layers plus one for convolutional
models, sequence length for recurrent models). Times are single-run wall-clock
measurements in milliseconds. Energy values are given with three significant
figures.
Figure~\ref{fig:timeline} in Section~\ref{sec:exp-efficiency} draws one CIFAR-10
step of this schedule to scale.

A few exceptions are visible: five datasets (SVHN, BullyDetect, Widar3,
AOPHand, and MMActivity) have optimized time ratios above one, reaching $2.11$
on BullyDetect. Apart from SVHN, which is essentially at parity ($1.004$),
these are shallow or small-scale models that offer few parallel gradient tasks
(small $P$), so the projected schedule cannot amortize the additional local
losses. These exceptions should be read in context: the energy ratio is below
one on \emph{all} $25$ datasets, the time ratio exceeds one on only $5$ of
$25$, and the reductions achieved elsewhere are much larger than these
increases (down to $0.04\times$ for energy and $0.29\times$ for time on
NTU-Fi-HumanID). Because the ratios are multiplicative quantities, we summarize
them with the geometric mean, $0.36\times$ for energy and $0.72\times$ for
optimized time across all datasets, so BSD remains clearly cheaper than BP in
aggregate. We further note that our models use generic multi-layer CNN, LSTM,
and RNN architectures whose hyperparameters were tuned for accuracy rather than
training cost. For the datasets with ratios above one, adjusting the
architecture and efficiency-related hyperparameters (e.g., layer widths,
alignment-point placement, or segment length) offers clear room to reduce the
local-gradient overhead, which we leave to future work. Further limitations are
discussed in Appendix~\ref{app:limitations}.

\begin{table*}[h]
  \centering
  \footnotesize
  \setlength{\tabcolsep}{4.5pt}
  \vspace{-10pt}
  \caption{Per-sample operation counts and estimated training energy of the
  matched BP and BSD models. MACs and ACs are given in millions and energy in
  mJ, all with three significant figures.}
  \label{tab:energy-full}
  \begin{tabular}{@{}ll rr rr rr r@{}}
    \toprule
    & & \multicolumn{2}{c}{MACs ($\times10^{6}$)} & \multicolumn{2}{c}{ACs ($\times10^{6}$)}
    & \multicolumn{2}{c}{Energy (mJ)} & \multicolumn{1}{c@{}}{Ratio} \\
    \cmidrule(lr){3-4}\cmidrule(lr){5-6}\cmidrule(lr){7-8}\cmidrule(l){9-9}
    Modality & Dataset & BP & BSD & BP & BSD & BP & BSD & \multicolumn{1}{c@{}}{BSD/BP} \\
    \midrule
    \multirow{6}{*}{Vision} & MNIST & 106 & 27.2 & 45.9 & 63.8 & 0.528 & 0.182 & 0.345 \\
     & Fashion-MNIST & 122 & 27.2 & 53.9 & 76.1 & 0.609 & 0.194 & 0.318 \\
     & CIFAR-10 & 158 & 36.8 & 57.9 & 78.2 & 0.780 & 0.240 & 0.307 \\
     & CIFAR-100 & 157 & 36.8 & 57.4 & 76.0 & 0.775 & 0.238 & 0.307 \\
     & SVHN & 150 & 36.8 & 53.8 & 70.6 & 0.738 & 0.233 & 0.315 \\
     & Tiny-ImageNet & 650 & 147 & 240 & 315 & 3.21 & 0.961 & 0.300 \\
    \midrule
    \multirow{4}{*}{Motion} & UCI-HAR & 12.8 & 7.37 & 2.96 & 9.55 & 0.0617 & 0.0425 & 0.689 \\
     & HHAR & 137 & 133 & 2.77 & 9.90 & 0.632 & 0.619 & 0.979 \\
     & MotionSense & 13.9 & 8.29 & 2.81 & 9.55 & 0.0666 & 0.0467 & 0.702 \\
     & Shoaib & 91.3 & 71.5 & 1.16 & 4.55 & 0.421 & 0.333 & 0.791 \\
    \midrule
    \multirow{4}{*}{Acoustic} & ESC-50 & 494 & 110 & 219 & 323 & 2.47 & 0.795 & 0.322 \\
     & UrbanSound8K & 483 & 110 & 213 & 311 & 2.42 & 0.784 & 0.325 \\
     & GSC & 370 & 206 & 10.9 & 34.7 & 1.71 & 0.981 & 0.573 \\
     & GTZAN & 613 & 110 & 278 & 370 & 3.07 & 0.838 & 0.273 \\
    \midrule
    \multirow{8}{*}{Wireless} & UT-HAR & 4{,}020 & 176 & 1{,}850 & 2{,}980 & 20.2 & 3.49 & 0.173 \\
     & NTU-Fi-HumanID & 14{,}900 & 470 & 6{,}280 & 1{,}050 & 74.3 & 3.10 & 0.042 \\
     & NTU-HAR & 4{,}150 & 491 & 1{,}490 & 2{,}240 & 20.4 & 4.28 & 0.209 \\
     & Widar3 & 192 & 128 & 18.2 & 58.6 & 0.900 & 0.642 & 0.713 \\
     & ARIL & 1{,}760 & 147 & 813 & 1{,}400 & 8.85 & 1.93 & 0.218 \\
     & BullyDetect & 1{,}060 & 304 & 374 & 1{,}940 & 5.21 & 3.14 & 0.603 \\
     & AOPHand & 16.8 & 7.60 & 5.94 & 15.9 & 0.0826 & 0.0492 & 0.596 \\
     & MMActivity & 74.4 & 28.2 & 7.71 & 14.3 & 0.349 & 0.143 & 0.408 \\
    \midrule
    \multirow{3}{*}{Neuromorphic} & CIFAR10-DVS & 897 & 98.4 & 369 & 642 & 4.46 & 1.03 & 0.231 \\
     & DVS-Gesture & 1{,}310 & 229 & 373 & 990 & 6.37 & 1.94 & 0.305 \\
     & SHD & 283 & 85.3 & 120 & 185 & 1.41 & 0.559 & 0.397 \\
    \bottomrule
  \end{tabular}
\end{table*}
\begin{table*}[h]
  \vspace{-10pt}
  \centering
  \scriptsize
  \setlength{\tabcolsep}{3.8pt}
  \vspace{-20pt}
  \caption{Profiled phase times per training step (ms) and the resulting step
  times, $T_{\mathrm{opt}}(\mathrm{BP})=T_F+T_L+T_U$ and
  $T_{\mathrm{opt}}(\mathrm{BSD})=\max(T_F,T_R)+(T_L+T_U)/P$. $T_R$ applies
  only to BSD, since BP has no reverse branch.}
  \label{tab:time-full}
  \begin{tabular}{@{}ll rr r rr rr rr r@{}}
    \toprule
    & & \multicolumn{2}{c}{$T_F$} & \multicolumn{1}{c}{$T_R$} & \multicolumn{2}{c}{$T_L$}
    & \multicolumn{2}{c}{$T_U$} & \multicolumn{2}{c}{$T_{\mathrm{opt}}$} & \multicolumn{1}{c@{}}{Ratio} \\
    \cmidrule(lr){3-4}\cmidrule(lr){5-5}\cmidrule(lr){6-7}\cmidrule(lr){8-9}\cmidrule(lr){10-11}\cmidrule(l){12-12}
    Modality & Dataset & BP & BSD & BSD & BP & BSD & BP & BSD & BP & BSD & \multicolumn{1}{c@{}}{BSD/BP} \\
    \midrule
    \multirow{6}{*}{Vision}
    & MNIST & 16.0 & 21.6 & 21.9 & 17.7 & 62.6 & 2.08 & 2.63 & 35.8 & 32.8 & 0.915 \\
    & Fashion-MNIST & 18.8 & 14.8 & 20.6 & 18.8 & 64.0 & 2.09 & 2.48 & 39.7 & 31.7 & 0.798 \\
    & CIFAR-10 & 14.9 & 12.5 & 14.1 & 21.2 & 71.2 & 1.76 & 2.42 & 37.8 & 26.4 & 0.697 \\
    & CIFAR-100 & 16.4 & 16.1 & 17.8 & 23.7 & 77.8 & 1.84 & 3.06 & 41.9 & 31.3 & 0.747 \\
    & SVHN & 12.4 & 12.9 & 14.8 & 17.3 & 99.4 & 2.01 & 2.68 & 31.7 & 31.8 & 1.004 \\
    & Tiny-ImageNet & 24.9 & 29.8 & 24.6 & 48.5 & 113.1 & 2.74 & 2.58 & 76.1 & 49.1 & 0.645 \\
    \midrule
    \multirow{4}{*}{Motion}
    & UCI-HAR & 135.7 & 226.8 & 151.7 & 261.7 & 744.6 & 1.75 & 3.49 & 399.2 & 232.7 & 0.583 \\
    & HHAR & 85.6 & 103.3 & 111.0 & 83.8 & 425.0 & 1.28 & 2.76 & 170.7 & 114.3 & 0.670 \\
    & MotionSense & 182.8 & 154.1 & 188.6 & 184.2 & 717.3 & 1.78 & 2.98 & 368.7 & 194.2 & 0.527 \\
    & Shoaib & 40.4 & 49.3 & 53.4 & 40.0 & 198.0 & 1.06 & 2.34 & 81.4 & 56.8 & 0.697 \\
    \midrule
    \multirow{4}{*}{Acoustic}
    & ESC-50 & 26.6 & 20.5 & 23.8 & 49.3 & 94.5 & 1.52 & 2.17 & 77.4 & 43.1 & 0.557 \\
    & UrbanSound8K & 22.7 & 25.1 & 14.7 & 46.0 & 86.4 & 3.29 & 4.65 & 72.0 & 43.3 & 0.601 \\
    & GSC & 58.6 & 70.9 & 71.0 & 70.6 & 315.1 & 0.99 & 2.29 & 130.2 & 74.1 & 0.569 \\
    & GTZAN & 17.7 & 33.8 & 27.1 & 42.6 & 93.4 & 1.70 & 2.84 & 62.0 & 53.1 & 0.856 \\
    \midrule
    \multirow{8}{*}{Wireless}
    & UT-HAR & 16.5 & 24.3 & 19.5 & 39.0 & 82.6 & 1.93 & 2.44 & 57.5 & 41.3 & 0.718 \\
    & NTU-Fi-HumanID & 42.5 & 19.1 & 17.5 & 91.5 & 80.8 & 2.75 & 2.17 & 136.7 & 39.8 & 0.291 \\
    & NTU-HAR & 23.7 & 25.6 & 26.0 & 45.6 & 69.9 & 2.59 & 3.50 & 71.9 & 44.3 & 0.616 \\
    & Widar3 & 7.6 & 16.2 & 16.8 & 12.5 & 73.2 & 1.49 & 3.04 & 21.6 & 32.1 & 1.486 \\
    & ARIL & 23.7 & 26.2 & 20.1 & 51.4 & 83.8 & 4.48 & 1.80 & 79.6 & 47.7 & 0.598 \\
    & BullyDetect & 12.5 & 13.9 & 35.1 & 18.4 & 130.5 & 1.55 & 2.31 & 32.4 & 68.3 & 2.110 \\
    & AOPHand & 10.4 & 12.7 & 14.2 & 11.8 & 44.9 & 2.39 & 2.65 & 24.6 & 26.1 & 1.063 \\
    & MMActivity & 10.6 & 13.9 & 15.3 & 14.1 & 54.4 & 1.38 & 2.33 & 26.1 & 26.6 & 1.020 \\
    \midrule
    \multirow{3}{*}{Neuromorphic}
    & CIFAR10-DVS & 22.6 & 26.0 & 17.9 & 73.3 & 71.9 & 0.85 & 1.29 & 96.7 & 44.3 & 0.458 \\
    & DVS-Gesture & 41.6 & 29.9 & 36.3 & 71.6 & 99.0 & 1.98 & 4.03 & 115.2 & 56.9 & 0.494 \\
    & SHD & 14.1 & 15.9 & 17.9 & 38.3 & 87.2 & 1.49 & 1.84 & 53.9 & 35.7 & 0.663 \\
    \bottomrule
  \end{tabular}
  \vspace{-10pt}
\end{table*}
\section{Batch-Size Sensitivity}
\label{app:batch-sensitivity}

The off-diagonal term of the ReCo objective relies on negative sample pairs
within a batch, so the contrastive signal weakens as the batch shrinks. Since
edge devices may train with small batches under tight memory budgets, we
measure how BSD accuracy varies with batch size. We evaluate one
representative dataset per modality (CIFAR-10, MotionSense, GTZAN, AOPHand,
and SHD), training the BSD model with its per-dataset configuration from
Table~\ref{tab:dataset-configs} and varying only the batch size over
$\{64, 32, 16, 8, 4\}$.

\begin{table}[h]
  \vspace{-10pt}
  \centering
  \small
  \setlength{\tabcolsep}{7pt}
  \caption{BSD test accuracy (\%) under different training batch sizes.
  Performance is largely preserved down to a batch size of $32$ and degrades
  noticeably below $16$.}
  \label{tab:batch-sensitivity}
  \begin{tabular}{@{}ll ccccc@{}}
    \toprule
    & & \multicolumn{5}{c@{}}{Training batch size} \\
    \cmidrule(l){3-7}
    Modality & Dataset & 64 & 32 & 16 & 8 & 4 \\
    \midrule
    Vision & CIFAR-10 & 81.5 & 78.1 & 69.0 & 50.0 & 36.1 \\
    Motion & MotionSense & 95.7 & 95.1 & 94.5 & 90.5 & 79.4 \\
    Acoustic & GTZAN & 61.5 & 60.5 & 60.0 & 51.0 & 43.5 \\
    Wireless & AOPHand & 80.6 & 76.8 & 75.8 & 71.2 & 60.1 \\
    Neuromorphic & SHD & 98.5 & 96.5 & 93.9 & 84.9 & 83.0 \\
    \bottomrule
  \end{tabular}
\end{table}

Table~\ref{tab:batch-sensitivity} shows that accuracy is largely preserved
down to a batch size of $32$ and degrades noticeably below $16$, confirming
that ReCo needs a sufficient number of negative pairs per batch. Two
observations put this requirement in context. First, the batch dimension is
generally much smaller than the feature dimensions of the network, so the
additional activation memory incurred by using a batch of $32$ instead of a
very small one is limited. Second, a batch of $32$ is already relatively
small for on-device training, and the performance at this size---at most a few
points below the large-batch setting---remains acceptable.

\section{Limitations and Future Directions}
\label{app:limitations}

Our efficiency results are operation-level estimates and timing profiles
collected on general-purpose GPUs; the models have not yet been deployed on
actual edge devices. Validating the measured and projected costs on real
hardware is therefore an important next step. In particular, the projected
training-time reduction assumes that the two branches and the local-gradient
tasks can execute concurrently, so approaching the theoretical optimum may
require specialized compute devices or neuromorphic chips that expose this
parallelism.

Second, the architectures to which BSD has been applied are still relatively
limited. Our evaluation covers convolutional, recurrent, and gated-recurrent
models, and it remains to be explored whether the same bidirectional
construction can train larger families such as Transformers, where the
attention-based reverse branch and its alignment points require a dedicated
design.

\section{The Use of Large Language Models}
\label{app:llm}

Large language models were used to assist with literature retrieval, text
editing, and the preparation of plotting scripts during the writing of this
paper. The authors reviewed, verified, and edited all generated content and
take full responsibility for the final manuscript, including all reported
claims, data, and citations.

\end{document}